\documentclass[
]{ceurart}

\usepackage{caption}
\usepackage{subcaption}
\usepackage{listings}
\usepackage{xspace}
\usepackage{amssymb}
\usepackage{amsmath}
\usepackage{xcolor}
\usepackage{pgfplots}
\usepackage{pgfplotstable}
\usepackage{tikz}
\usepackage{enumitem}
\usepackage{footnote}
\makesavenoteenv{tabular}
\makesavenoteenv{table}
\usepackage{algorithm}
\usepackage{algpseudocode}

\definecolor{gray}{HTML}{B8B8B8}
\definecolor{TEAL}{HTML}{298C8C}
\definecolor{MAGENTA}{HTML}{800074}
\definecolor{bblue}{HTML}{4F81BD}
\definecolor{rred}{HTML}{C0504D}
\definecolor{ggreen}{HTML}{9BBB59}
\definecolor{ppurple}{HTML}{9F4C7C}
\definecolor{oorange}{HTML}{9F7F4C}

\usepackage[misc,geometry]{ifsym}
\usepackage{xspace}
\usepackage{amsmath}
\newcommand{\ie}{i.e.,\xspace}

\newcommand{\st}{s.t.,\xspace} 
\newcommand{\eg}{e.g.,\xspace}

\newcommand{\cf}{cf.\xspace}
\newcommand{\etc}{etc.\xspace}
\newcommand{\qq}[1]{``#1''}

\newcommand{\gtriuparrow}{\textcolor{green}{$\blacktriangle$}}

\newcommand{\mKGQAgent}{mKGQAgent\xspace}
\newcommand{\SimpleAgent}{$\mathcal{SA}gent$\xspace}

\newcommand{\Fscore}{F1 score\xspace}

\newcommand{\LLM}{LLM\xspace}

\newcommand{\LLMs}{LLMs\xspace}

\newcommand{\QA}{QA\xspace}

\newcommand{\KGQA}{KGQA\xspace}

\newcommand{\KG}{KG\xspace}

\newcommand{\QALDplus}{QALD-9-plus\xspace}

\newcommand{\SPARQL}{SPARQL\xspace}

\newcommand{\Wikidata}{Wikidata\xspace}

\newcommand{\RQ}[1]{$\mathcal{RQ}$\textit{#1}}

\newcommand{\gtriuparrowred}{\textcolor{red}{$\blacktriangle$}}
\newcommand{\gtridownarrowred}{\textcolor{red}{$\blacktriangledown$}}
\begin{document}


\conference{First International TEXT2SPARQL Challenge Co-Located with Text2KG at ESWC 2025}

\title{Text-to-SPARQL Goes Beyond English: Multilingual Question Answering Over Knowledge Graphs through Human-Inspired Reasoning}


\author[1,2]{Aleksandr Perevalov}[%
orcid=0000-0002-6803-3357,
email=aleksandr.perevalov@htwk-leipzig.de,
url=https://perevalov.com,
]
\cormark[1]
\address[1]{WSE Research Group, Leipzig University of Applied Sciences,
  Karl-Liebknecht-Straße 132, 04277, Leipzig, Germany}
\address[2]{DICE Research Group, University of Paderborn,
  Warburger Str. 100, 33098, Paderborn, Germany}

\author[1]{Andreas Both}[%
orcid=0000-0002-9177-5463,
email=andreas.both@htwk-leipzig.de,
url=http://andreasboth.de,
]

\cortext[1]{Corresponding author.}

\begin{abstract}
Accessing knowledge via multilingual natural-language interfaces is one of the emerging challenges in the field of information retrieval and related ones.
Structured knowledge stored in knowledge graphs can be queried via a specific query language (e.g., SPARQL). 
Therefore, one needs to transform natural-language input into a query to fulfill an information need.
Prior approaches mostly focused on combining components (e.g., rule-based or neural-based) that solve downstream tasks and come up with an answer at the end.
We introduce mKGQAgent, a human-inspired framework that breaks down the task of converting natural language questions into SPARQL queries into modular, interpretable subtasks. 
By leveraging a coordinated LLM agent workflow for planning, entity linking, and query refinement---guided by an experience pool for in-context learning---mKGQAgent efficiently handles multilingual KGQA. 
Evaluated on the DBpedia- and Corporate-based KGQA benchmarks within the Text2SPARQL challenge 2025, our approach took first place among the other participants.
This work opens new avenues for developing human-like reasoning systems in multilingual semantic parsing.
\end{abstract}


\maketitle

\section{Introduction}

Previous approaches to multilingual knowledge graph question answering (\KGQA), like  \citet{qanswer,deeppavlov2023}, have employed both rule-based and neural methods to address downstream tasks (\eg named entity recognition, relation detection, query template classification) necessary for constructing structured queries (\eg \SPARQL queries).
More recent methods (\eg \citet{srivastava2024mst5}) leverage Large Language Models (\LLMs) to generate such structured queries directly from non-English input. 
The application of newly introduced \emph{\LLM agents} (or \emph{augmented} language models) to \KGQA has demonstrated significantly improved performance compared to \LLMs that rely solely on standard prompting techniques \eg \citet{jiang2024kg,huang2024queryagent}). 
However, the multilingual aspect of these systems remains largely unexplored within the research community.
To the best of our knowledge, \emph{there are no studies investigating the \LLM agent architectures for \KGQA in multilingual settings}. 

One of the key advantages of \LLMs is that they enable developers and researchers to model human-like reasoning processes via agentic workflows (\cf \citet{li2023metaagents}). 
When solving complex problems, humans typically break them down into a series of simpler subtasks (\cf  \citet{diefenbach2017qanaryecosystem,correa2020resource}), effectively creating a step-by-step \emph{plan} to arrive at a solution.
While generating a \SPARQL query, this decomposition is essential: not only does one need to break down the task, but also \emph{look up} query language syntax, identify relevant entity identifiers in the target knowledge graph (\KG), and analyze \emph{feedback} (\eg from executing the SPARQL query candidate on the triplestore).
To replicate this human-like process, we introduce \mKGQAgent--an \LLM-based agent framework designed as a \KGQA system that follows a semantic-parsing approach. 
Specifically, given a user query (multiple languages are supported), it generates a \SPARQL query to fulfill the information need.
Accordingly, this paper aims to answer the following \emph{research questions}:
\begin{description}[labelindent=-2pt]
    \item[\RQ{1}] How do different LLM agent steps (\eg plan, action, tool calling, feedback, \etc) impact the generation of \SPARQL queries from natural language?
    \item[\RQ{2}] How efficient are these LLM agent steps in terms of computation time and the number of additional calls required?
    \item[\RQ{3}] How does the quality of \SPARQL query generation vary when prompting LLM agents in non-English languages (especially low-resource ones)?
    \item[\RQ{4}] How does translating non-English questions into English affect the quality of KGQA?
\end{description}

We conducted preliminary experiments on the widely used KGQA benchmark \QALDplus (introduced in \citet{perevalov2022qald}) with multilingual support. 
We evaluate 10 languages, including two classified as endangered. 
The experimental results on both proprietary and open-source LLMs demonstrate the effectiveness of \mKGQAgent's architecture, achieving superior performance even in non-English settings.
During the final evaluation on the DBpedia- and Corporate-based KGQA benchmarks within the Text2SPARQL challenge 2025, our approach took first place among the other participants.
The source code and the evaluation results are available in our GitHub repository\footnote{\url{https://github.com/WSE-research/text2sparql-agent}}.

The paper is organized as follows. 
In the next section, an overview of the related work is presented.
The \mKGQAgent architecture is described in Section \ref{sec:kgqagent_architecture}.
Section \ref{sec:experimental_setup} is dedicated to the experimental setup.
The results are shown in Section \ref{sec:results_insights} and discussed in Section \ref{sec:discussion}.
Section \ref{sec:conclusion} concludes our paper.

\section{Related Work}\label{sec:related_work}

Recent KGQA research has included classical, rule-based, and neural approaches \cite{leaderboard,perevalovmultilingual}.
\citet{qanswer} (QAnswer) and \citet{10.1145/3281354.3281362} used query templates and rule indexes without language models.
\citet{platypus} applied grammar rules for \SPARQL query transformation.
DeepPavlov 2023 employs a fine-tuned language model pipeline for query generation, \cf \citet{deeppavlov2023}.
\citet{omar2023universal} proposed KGQAN, which integrates answer type prediction and triple pattern generation.

Multilingual KGQA solutions including \citet{zhou-etal-2021-improving}, which fine-tune multilingual transformers and leverage bilingual lexicon induction.
\citet{zhang-etal-2023-xsemplr} address cross-lingual semantic parsing over multiple meaning representations in XSemPLR, including \SPARQL.
\citet{TAN2023120721} improve cross-lingual reasoning, enhancing the Entity Alignment model performance in \emph{English, Chinese, and French} in the CLRN approach.

\citet{zong2024triad} employ the multi-role LLM agent architecture Triad for \SPARQL query construction.
MST5 (\citet{srivastava2024mst5}) fine-tunes mT5-XL for generating structured queries.
\citet{lehmann2024beyond} enhance LLMs with external tools to mimic human-like reasoning.
\citet{jiang2024kg} integrates a KG-based executor (KG-Agent) and fine-tunes {Llama2}-7B for improved tool usage.
QueryAgent (\citet{huang2024queryagent}) mitigates hallucinations with ERASER-based self-correction, excelling on GrailQA and GraphQ.
Interactive-KBQA (\citet{xiong2024interactive}) iteratively refines LLM outputs via direct KB interactions.

\section{The \mKGQAgent Architecture}\label{sec:kgqagent_architecture}
\begin{figure}[h!]
  \centering
    \frame{\includegraphics[width=\linewidth]{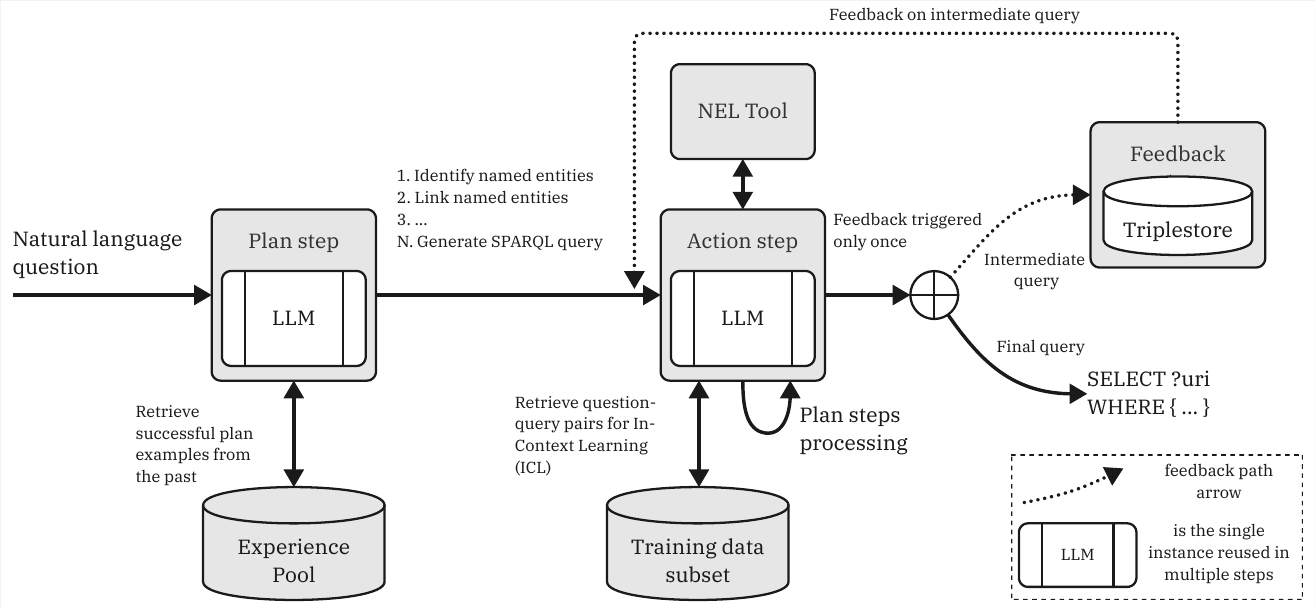}}
  \caption{
  \mKGQAgent workflow demonstration (online phase). In the evaluation phase, the \mKGQAgent is using the experience pool examples to improve planning, the in-context learning training examples to improve \SPARQL query generation awareness and the feedback to correct possible errors.
   The offline phase, which is required for gathering experience pool. The evaluation or online phase -- the routing of the \mKGQAgent's components as well as their integral modules.}
  \label{fig:kgqagent-workflow}
\end{figure}
\begin{figure}[h!]
    \centering
    \frame{\includegraphics[width=\linewidth]{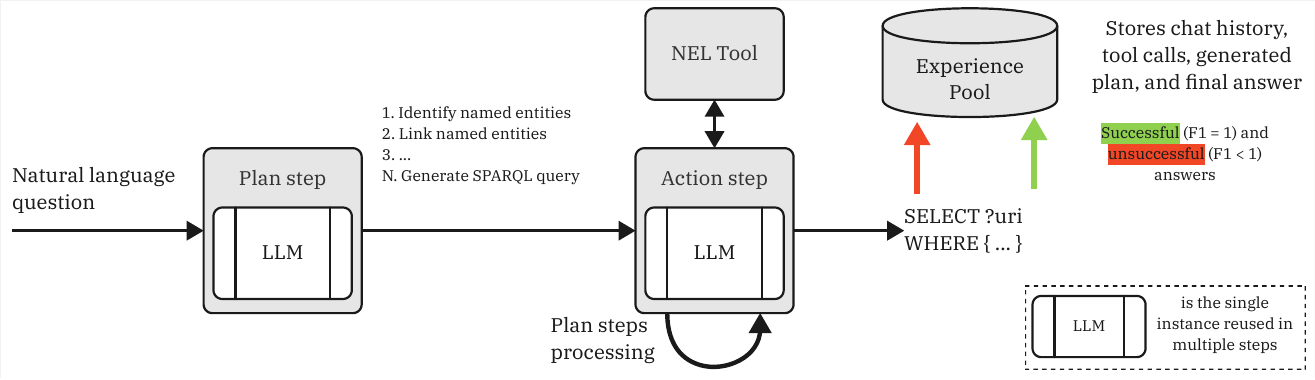}}
    \caption{\SimpleAgent workflow demonstration (offline phase). In the offline phase, the \SimpleAgent is used to gather the experience pool of positive and negative answers over the train subsets.}
    \label{fig:simple-agent-workflow}
  \end{figure}

The \mKGQAgent workflow consists of several key steps (see Figure \ref{fig:kgqagent-workflow} for an overview).
Our approach follows the terminology established in recent survey articles on \LLM agents, \cf \citet{mialon2023augmented,wang2024survey}. 
The framework operates in two main phases: the \emph{offline phase} and the \emph{evaluation (online) phase}.
The offline phase is essential for preparing the \emph{experience pool} (see Section \ref{ssec:experience-pool-construction}).
During the offline phase, we employ the \emph{simple agent} (\SimpleAgent) to gather intermediate processing steps for the experience pool (see Figure \ref{fig:simple-agent-workflow}). 

\SimpleAgent uses the \emph{plan step} (\cf Section~\ref{ssec:plan-simple}) to generate a structured step-by-step plan and the \emph{action step}, that either calls the LLM or the \emph{named entity linking (NEL) tool} (\cf Section~\ref{ssec:nel-tool}) ultimately leading to the \SPARQL query generation.
In the \emph{evaluation (online) phase}, the \mKGQAgent is using the plan step and the action step with the experience pool and the NEL tool, and the feedback step that has access to the triplestore.

The important feature of our framework is that it does not require supervised fine-tuning, which significantly reduces the computation costs and preserves the generalizability of the original \LLMs (\cf catastrophic forgetting); see \citet{luo2023empirical}.

\subsection{Offline Phase of the \mKGQAgent}
\subsubsection{Named Entity Linking (NEL) Tool}\label{ssec:nel-tool}

Likewise, humans look up a resource identifier in a KG, and the NEL step interacts with the environment (\ie KG) and retrieves resource labels from there.
Assuming the fact that an LLM was not given the URI-label mappings of a particular resource, the \SPARQL query generation would not be possible.
Importantly, while introducing the NEL tool, we do not propose a novel NEL algorithm.
In contrast, we demonstrate how to utilize an existing NEL service in the LLM agent workflow (see Algorithm \ref{alg:entity-linking}).

\begin{algorithm}[h!]
\caption{NEL Tool}
\label{alg:entity-linking}
\begin{algorithmic}[1]
\Require Entity candidates $E$, Relation candidates $R$, NEL service $\mathcal{NEL}$
\Ensure Dictionary with linked entities: linkedEntities
\Ensure Dictionary with linked relations: linkedRelations
\State Initialize empty dictionaries linkedEntities and linkedRelations
\For{each entity $e_i \in E$}
    \State $e^{\text{URI}}_i \gets \mathcal{NEL}(e_i)$
    \If{$e^{\text{URI}}_i$ is not empty}
        \State $\text{linkedEntities}[e_i] \gets e^{\text{URI}}_i$
    \EndIf
\EndFor
\For{each relation $r_j \in R$}
    \State $r^{\text{URI}}_j \gets \mathcal{NEL}(e_j)$
    \If{$r^{\text{URI}}_j$ is not empty}
        \State $\text{linkedEntities}[r_j] \gets r^{\text{URI}}_j$
    \EndIf
\EndFor
\State \Return (linkedEntities, linkedRelations)
\end{algorithmic}
\end{algorithm}

The entity and relation candidates are proposed by the backbone LLM within the tool calling process at the action step (see Sections \ref{ssec:action-step-simple} and \ref{ssec:action-step-experience}).
Entity and relation linking is crucial for the text-to-\SPARQL process since the URIs representing resources in a KG may be done using random identifiers\footnote{\eg in Wikidata \citet{vrandevcic2014wikidata}, \href{https://www.wikidata.org/wiki/Q567}{Q567} (\url{https://www.wikidata.org/wiki/Q567}) for \qq{Angela Merkel}}.

\subsubsection{Plan step}\label{ssec:plan-simple}
The plan step leverages the backbone LLM to generate a step-by-step list of tasks to come up with a \SPARQL query given a question.
The intuition behind the plan step is that it simplifies the task for the model such that it does not need to handle the whole complexity at once.
For example, such tasks as entity recognition and linking, query refinement, etc.
Thus, following the human-like behavior (\cf \citet{huys2015interplay,correa2020resource}), the plan step intends to break down the complex task of writing a SPARQL query into a combination of simpler subtasks.
Hence, the action step deals at one point in time with a simple subtask having the results of the previous steps in its conversation history.
For details regarding the plan step (for details, see Algorithm \ref{alg:plan-step}).

\begin{algorithm}[h!]
\caption{Plan step without experience pool}
\label{alg:plan-step}
\begin{algorithmic}[1]
\Require Natural language question $q_i$, system prompt $S_{\text{plan}}$, model $\mathcal{LLM}$
\Ensure Step-by-step plan $p_i$ \Comment{List of textual tasks}
\State $p_i \gets \mathcal{LLM}(S_{\text{plan}}, q_i)$ \Comment{Query LLM with system prompt and question}
\State \Return $p_i$
\end{algorithmic}
\end{algorithm}

\subsubsection{Action Step without Experience Pool}\label{ssec:action-step-simple}

Once the plan is generated, the action step executes each of the plan tasks sequentially, leveraging the NEL Tool for the entity linking (see Algorithm \ref{alg:action-step}).
This approach ensures that the agent follows the structured plan, interacting with necessary tools to refine and complete the \SPARQL query generation process.

\begin{algorithm}[h!]
\caption{Action Step without Experience Pool}
\label{alg:action-step}
\begin{algorithmic}[1]
\Require Step-by-step plan $p_i$, model $\mathcal{LLM}$, tool $NEL$, system prompt $S_{\text{action}}$
\Ensure Generated SPARQL query $\hat{\phi_i}$
\State Bind $NEL$ to $\mathcal{LLM}$
\State Initialize empty chat history $\mathcal{H}_i$
\For{each step $s_j \in p_i$}
\State $h_j \gets \mathcal{LLM}(S_{\text{action}}, s_j)$ \Comment{LLM may call tool or just itself}
\State Append $h_j$ to $\mathcal{H}_i$
\EndFor
\State $\hat{\phi_i} \gets \text{lastElementOf}(\mathcal{H}_i)$
\State \Return $\hat{\phi_i}$
\end{algorithmic}
\end{algorithm}

\subsubsection{Experience Pool Construction}\label{ssec:experience-pool-construction}
During the offline phase, we utilize \SimpleAgent to collect the \emph{experience pool}. 
This involves evaluating the correctness of the generated \SPARQL queries (based on the ground truth data) and storing them together with the intermediate steps (i.e., plan, chat history) in a vector database (see Algorithm \ref{alg:experience-pool-construction}). 
Therefore, each natural language question from the train subset is converted into a vector representation that is associated with metadata, including the corresponding plan, intermediate steps of the action step, and the final results.
The experience pool is a non-parametric memory of our agent that contains both successful (\Fscore $=1.0$) and unsuccessful (\Fscore $<1.0$) \SPARQL query generation attempts based on a ground truth.

\begin{algorithm}[t]
\caption{Add Example to the Experience Pool}
\label{alg:experience-pool-construction}
\begin{algorithmic}[1]
\Require Training set example $d_i \in \mathcal{D}_{\text{train}}$, step-by-step plan $p_i$, chat history $\mathcal{H}_i$, Experience pool $\mathcal{E}$, Text embedding model $\mathcal{EMB}$
\Ensure Updated experience pool $\mathcal{E}'$
\State $q_i, \phi_i \gets d_i$ \Comment{Unpack training example (question and ground truth \SPARQL)}
\State $\hat{\phi_i} \gets \text{lastElementOf}(\mathcal{H})$ \Comment{Get the \SPARQL generated by \SimpleAgent}
\State  $\text{F1}_i \gets F1_{\text{score}}(\phi_i, \hat{\phi_i})$ \Comment{Compute F1 score}
\State $v_{q_i} \gets \mathcal{EMB}(q_i)$ \Comment{Convert question to a vector}
\State $\mathcal{E}' \gets \mathcal{E} + \{q_i, v_{q_i}, \phi_i, \hat{\phi_i}, p_i, \mathcal{H}, \text{F1}_i\}$
\State \Return $\mathcal{E}'$
\end{algorithmic}
\end{algorithm}

Therefore, the experience pool holds the information about the quality of the generated \SPARQL queries ($\text{F1}_i$), the step-by-step plan ($p_i$) that was used to generate this particular query, and other metadata (\eg ground truth \SPARQL query).

\subsection{Evaluation Phase of the \mKGQAgent}\label{ssec:evaluation-phase}
\subsubsection{Plan step with the Experience Pool}
In the evaluation phase, the plan step leverages the experience pool to find relevant plan examples for better planning quality.
The plan examples are included in the system prompt $S_\text{plan}$ (see Algorithm \ref{alg:plan-step-experience}).

Hence, the plan step benefits from the prior successful planning examples while using them in the system prompt for in-context learning.

\subsubsection{Action step with the Experience Pool}\label{ssec:action-step-experience}

Once the plan is generated, the action step executes each of the plan tasks sequentially, leveraging the NEL Tool for the entity linking (see Algorithm \ref{alg:action-step-experience}).
The usage of the experience pool ensures that the LLM benefits from the in-context \SPARQL query examples from the training subset.
It is important to note that the plan $p_i$ can be populated with the result of the feedback step (in case the feedback is triggered).

\subsubsection{Feedback Step}
%
\begin{algorithm}[t]
\caption{Plan step with Experience Pool}
\label{alg:plan-step-experience}
\begin{algorithmic}[1]
\Require Natural language question $q_i$, system prompt $S_{\text{plan}}$, model $\mathcal{LLM}$, experience pool $\mathcal{E}$, text embedding model $\mathcal{EMB}$
\Ensure Step-by-step plan $p_i$ \Comment{List of textual tasks}
\State $v_{q_i} \gets \mathcal{EMB}(q_i)$
\State $P \gets findTopNPlans(\mathcal{E}, v_{q_i})$ \Comment{Finds top-N similar plans with $\text{F1}=1.0$}
\State $S^{\text{experience}}_{\text{plan}} \gets S_{\text{plan}} + P$ \Comment{The plans are included to the prompt}
\State $p_i \gets \mathcal{LLM}(S^{experience}_{\text{plan}}, q_i)$ \Comment{Query LLM with system prompt and question}
\State \Return $p_i$
\end{algorithmic}
\end{algorithm}
\begin{algorithm}[t]
\caption{Action Step with the Experience Pool}
\label{alg:action-step-experience}
\begin{algorithmic}[1]
\Require Step-by-step plan $p_i$, model $\mathcal{LLM}$, tool $NEL$, system prompt $S_{\text{action}}$ (see appendix), experience pool $\mathcal{E}$, text embedding model $\mathcal{EMB}$
\Ensure Generated SPARQL query $\hat{\phi_i}$
\State Bind $NEL$ to $\mathcal{LLM}$
\State Initialize empty chat history $\mathcal{H}_i$
\State $v_{q_i} \gets \mathcal{EMB}(q_i)$
\State $P \gets findTopNQueries(\mathcal{E}, v_{q_i})$ \Comment{Finds top-N similar \SPARQL queries}
\State $S^{experience}_{\text{action}} \gets S_{\text{action}} + P$ \Comment{The queries are included to the prompt}

\For{each step $s_j \in p_i$}
\State $h_j \gets \mathcal{LLM}(S^{experience}_{\text{action}}, s_j)$ \Comment{LLM may call tool or just itself}
\State Append $h_j$ to $\mathcal{H}_i$
\EndFor
\State $\hat{\phi_i} \gets \text{lastElementOf}(\mathcal{H}_i)$
\State \Return $\hat{\phi_i}$
\end{algorithmic}
\end{algorithm}
\begin{algorithm}[t]
\caption{Feedback Step}
\label{alg:feedback-step}
\begin{algorithmic}[1]
\Require Intermediate query $\phi_i$, prompt template $S_{\text{feedback}}$ (see appendix), triplestore $\mathcal{KG}$
\Ensure Feedback prompt $S'_{\text{feedback}}$
\State $\mathcal{A}_i \gets \mathcal{KG}(\phi_i)$ \Comment{Query the triplestore and get the response}
\State $S'_{\text{feedback}} \gets S_{\text{feedback}} + \mathcal{A}_i $ \Comment{Populate prompt with the response}
\State \Return $S'_{\text{feedback}}$
\end{algorithmic}
\end{algorithm}

The feedback executes the generated \SPARQL query $\phi$ on a triplestore, collects the response, and integrates it into a pre-defined prompt template for the action step.
Once the first version of a \SPARQL query is generated (\ie the result of the last planning step executed at the action step), it is redirected to the feedback step.
The feedback is formulated only once per input question, \ie there are no multiple feedback options intended to avoid infinite loops.
The detailed feedback step workflow is defined in Algorithm \ref{alg:feedback-step}.
After that, the feedback $S'_{\text{feedback}}$ is redirected to the action step.
The action step executes the feedback to refine the \SPARQL query and delivers the final query as the result.
\section{Experimental Setup}\label{sec:experimental_setup}
We conduct our experiments on the commonly used KGQA benchmark: \QALDplus (\citet{perevalov2022qald}).
\QALDplus contains 558 questions in multiple languages and queries over DBpedia \cite{dbpedia} and Wikidata \cf \cite{Wikidata}.
We consider all available languages from \QALDplus, in addition, we also take the Spanish questions, which were contributed to this dataset later (\citet{soruco2023qald}).
The structure of \QALDplus includes question texts and the corresponding ground truth \SPARQL queries that return the expected answer to a question.
For the evaluation of \KGQA quality, we use the Macro F1 score \cite{usbeck2019benchmarking}.

\subsection{Large Language Models and Text Embedding Models}
In this work, we use both open-source and proprietary  LLMs.
The proprietary ones are provided by OpenAI\footnote{\url{https://platform.openai.com/docs/models}}, namely, GPT-3.5 (\texttt{gpt-3.5-turbo-0125}), and GPT-4o (\texttt{gpt-4o-2024-05-13}).
The models are accessed via the official Python SDK\footnote{\url{https://github.com/openai/openai-python}} with \texttt{temperature=0}, and other hyperparameters are set to default.

The open-source LLMs are: Qwen2.5 72B Instruct\footnote{\url{https://huggingface.co/Qwen/Qwen2-72B-Instruct-AWQ}} and Meta Llama 3.1 70B Instruct\footnote{\url{https://huggingface.co/hugging-quants/Meta-Llama-3.1-70B-Instruct-AWQ-INT4}}.
Both models were used with the AWQ (\citet{MLSYS2024_42a452cb}) quantization (4-bit) to fit into the memory.
The models were deployed via the vLLM framework (\citet{kwon2023efficient}).
The maximal context size of the models was set to 16384 tokens to avoid out-of-memory exceptions.
The other hyperparameters were set to default.
For the open-source LLMs, we used a server with two Nvidia L40S GPUs (each 48GB VRAM).

For creating text embeddings for the experience pool, we used a specific model trained for producing high-quality text embeddings for multilingual input -- \texttt{multilingual e5 large}\footnote{\url{https://huggingface.co/intfloat/multilingual-e5-large}} (introduced by \citet{wang2024multilingual}).
According to the MTEB leaderboard\footnote{\url{https://huggingface.co/spaces/mteb/leaderboard}} introduced by \citet{muennighoff2022mteb}, the model is listed among the top-3 in different languages (we considered embedding models with a size smaller than 1 billion parameters).

\subsection{Implementation of \mKGQAgent}

The \mKGQAgent \emph{architecture} is implemented within the LangChain framework\footnote{\url{https://python.langchain.com}} in Python.
This framework facilitates the integration of various components required for the agent's functionality.

The \emph{entity linking} within the NEL tool is implemented via \Wikidata's official public entity lookup endpoint\footnote{\url{https://www.wikidata.org/w/api.php?action=wbsearchentities}}.
This endpoint is capable of handling input in multiple languages.
The NEL tool also uses an \emph{external relation linker}, Falcon 2.0 (\citet{falcon2}), for enhanced linking capabilities.

The \emph{routing} between the plan, action, and feedback is implemented within the LangGraph framework\footnote{\url{https://langchain-ai.github.io/langgraph/}}, which is part of the LangChain ecosystem.

The \emph{prompts} used within the \mKGQAgent are written in different languages, \st they match the input question language. 
The prompts in English, German, and Russian were written by native speakers, the other prompts were acquired via machine translation and further structure validation.
We list the prompts in Figure \ref{fig:agent-prompts}.

\begin{figure}[h!]
    \centering
     \begin{subfigure}[b]{\textwidth}
         \centering
        \scriptsize


\begin{lstlisting}
You are an intelligent Knowledge Graph-based Question Answering system.

You can use the tools to help yourself only if you \
DON't have this information in chat history:
- 'wikidata_el' for named entity linking  \
(e.g. "Person name" -> "URI" or "is child of" -> "URI")
to determine URIs in the Wikidata KG

|\textcolor{blue}{\{QUESTION\_QUERY\_EXAMPLE\}} \textcolor{red}{\# comes from the experience pool}| 
\end{lstlisting}
         \vspace{-3.5ex}
         \caption{The system prompt ($S^{\text{experience}}_{action}$) for the action step with the usage of the experience pool $\mathcal{E}$\\[2ex]}
         \label{fig:system_prompt}
     \end{subfigure}
     \begin{subfigure}[b]{\textwidth}
         \centering
         \scriptsize

\begin{lstlisting}
For the given objective, come up with a simple step by step plan to write a SPARQL query. 
This plan should involve individual tasks (e.g., named entity linking, relation linking,
expected answer type classification), that if executed correctly \
will yield the correct SPARQL.
Do not add any superfluous steps. 
The result of the final step should be the final  SPARQL query. 
Don't propose to execute the query.
At the end step you MUST output exactly **ONE** SPARQL query string \
**without extra text or markdown**.

|\textcolor{blue}{\{USER\_QUESTION\}}|
|\textcolor{blue}{\{PLAN\_EXPERIENCE\_EXAMPLE\}} \textcolor{red}{\# comes from the experience pool}|
\end{lstlisting}

         \vspace{-3.5ex}
         \caption{The planning prompt ($S^{\text{experience}}_{\text{plan}}$) for the plan step with the usage of the experience pool $\mathcal{E}$\\[2ex]}
         \label{fig:planning_prompt}
     \end{subfigure}
     \vfill
     \begin{subfigure}[b]{\textwidth}
         \centering
         \scriptsize



\begin{lstlisting}[language=,escapeinside=||,frame=single]
This is feedback to your generated SPARQL query produced by executing it on a triplestore.
Please rework your query if neccessary.

Initial question: |\textcolor{blue}{\{USER\_QUESTION\}}|
Your query: |\textcolor{blue}{\{GENERATED\_SPARQL\}}\textcolor{red}{ \# intermediate SPARQL query}|

--- Start triplestore response ---
|\textcolor{blue}{\{FEEDBACK\}}\textcolor{red}{ \# comes from the query execution on a triplestore}|
--- End triplestore response ---

Make sure that the query is formatted correctly. 
No extra text. No markdown. Just plain SPARQL.
Determine whether to output a URI (SELECT ?uri), number (COUNT), date, \
boolean (ASK), string (SELECT ?label).
DON'T USE "SERVICE wikibase:label"
\end{lstlisting}
         \vspace{-3.5ex}
         \caption{The feedback prompt ($S_{\text{feedback}}$) for the feedback step\\[2ex]}
         \label{fig:feedback_prompt}
     \end{subfigure}
    \caption{The English versions of the prompts used within the \mKGQAgent. Placeholders are color-coded blue. Comments are color-coded red. Important: We use language-specific prompts for every considered language. In case of the evaluation within Text2SPARQL challenge, we used English prompts only despite the input language.}
    \label{fig:agent-prompts}
\end{figure}

The \SPARQL queries generated by the \mKGQAgent are executed on the official Wikidata \SPARQL endpoint\footnote{\url{https://query.wikidata.org/bigdata/namespace/wdq/sparql}}.

\subsection{Baselines}\label{ssec:baselines}
\newcommand{\itemx}[1][XXX]{\paragraph{#1}}
To compare the performance of the \mKGQAgent we select both \qq{pre-LLM era} KGQA systems and the ones that use different prompting techniques with LLMs.
Also, the baselines were selected in a way that they can generate \SPARQL queries over Wikidata.
In particular, the following approaches are selected for comparison with ours: QAnswer, Platypus, DeepPavlov 2023, KGQAN, Triad, MST5, and HQA (\cf Section \ref{sec:related_work}).

\paragraph{} The selection of the baselines was also influenced by the results reported in the KGQA leaderboard by \citet{leaderboard}.
We reuse the reported results in our paper for comparison with our \mKGQAgent approach.


\subsection{Machine Translation of the Input}

Following our research agenda \cite{perevalovMt,linguaFranca}, we evaluate how well machine translation to English serves as an alternative to processing non-English questions natively with the OPUS MT models; \cf \citet{helsinkiNLP}.



Our machine translation experiments are complementary to the main contribution and, therefore, are limited to the German, Russian, and Spanish languages.
We selected these languages as they all represent different language branches---the Germanic, Slavic, and Romance, respectively.
\section{Evaluation and Analysis}\label{sec:results_insights}
\subsection{English-only Comparison with the Baselines}
\begin{figure}[t]
\centering
\begin{tikzpicture}
\centering

    \begin{axis}[
        height=5cm,
        width=\linewidth,
        ymin=0,
        ymax=65,
        ytick={0,10,20,30,40,50,60}, 
        bar width=10pt,
        enlarge x limits={abs=0.3cm},
        ylabel={F1 score, \%},
        symbolic x coords={mKGQAgent (GPT-4o), HQA (GPT-4), QAnswer, KGQAN, HQA (GPT-3.5), MST5, mKGQAgent (Qwen 2.5 72B),Triad (GPT-4),DeepPavlov 2023,mKGQAgent (GPT-3.5),Triad (GPT-3.5),mKGQAgent (Llama 3.1 70B),Platypus},
        xtick={mKGQAgent (GPT-4o), HQA (GPT-4), QAnswer, KGQAN, HQA (GPT-3.5), MST5, mKGQAgent (Qwen 2.5 72B),Triad (GPT-4),DeepPavlov 2023,mKGQAgent (GPT-3.5),Triad (GPT-3.5),mKGQAgent (Llama 3.1 70B),Platypus},
        nodes near coords={\pgfmathprintnumber[fixed zerofill,precision=1]{\pgfplotspointmeta}},
        nodes near coords style={font=\fontsize{1}{10}\selectfont},
         xticklabel style={font=\scriptsize, rotate=45, anchor=east},
         yticklabel style={font=\scriptsize},
         ylabel style={font=\scriptsize}
    ]
    \addplot[ybar,fill=TEAL] coordinates {(mKGQAgent (GPT-4o),54.83)};
    \addplot[ybar,fill=gray] coordinates {(HQA (GPT-4), 50.00)};
    \addplot[ybar,fill=gray] coordinates {(QAnswer, 44.59)};
    \addplot[ybar,fill=gray] coordinates {(KGQAN, 44.07)};
    \addplot[ybar,fill=gray] coordinates {(HQA (GPT-3.5), 43.00)};
    \addplot[ybar,fill=gray] coordinates {(MST5, 41.87)};
    \addplot[ybar,fill=TEAL] coordinates {(mKGQAgent (Qwen 2.5 72B),41.77)};
    \addplot[ybar,fill=gray] coordinates {(Triad (GPT-4), 41.60)};
    \addplot[ybar,fill=gray] coordinates {(DeepPavlov 2023, 37.16)};
    \addplot[ybar,fill=TEAL] coordinates {(mKGQAgent (GPT-3.5), 33.85)};
    \addplot[ybar,fill=gray] coordinates {(Triad (GPT-3.5), 29.70)};
    \addplot[ybar,fill=TEAL] coordinates {(mKGQAgent (Llama 3.1 70B), 18.42)};
    \addplot[ybar,fill=gray] coordinates {(Platypus, 15.03)};

    \end{axis}
\end{tikzpicture}
\caption{Comparison between our \mKGQAgent approach (teal) and the baselines (grey) on English questions of \QALDplus.}
\label{fig:mkgqa-english-only}
\end{figure}
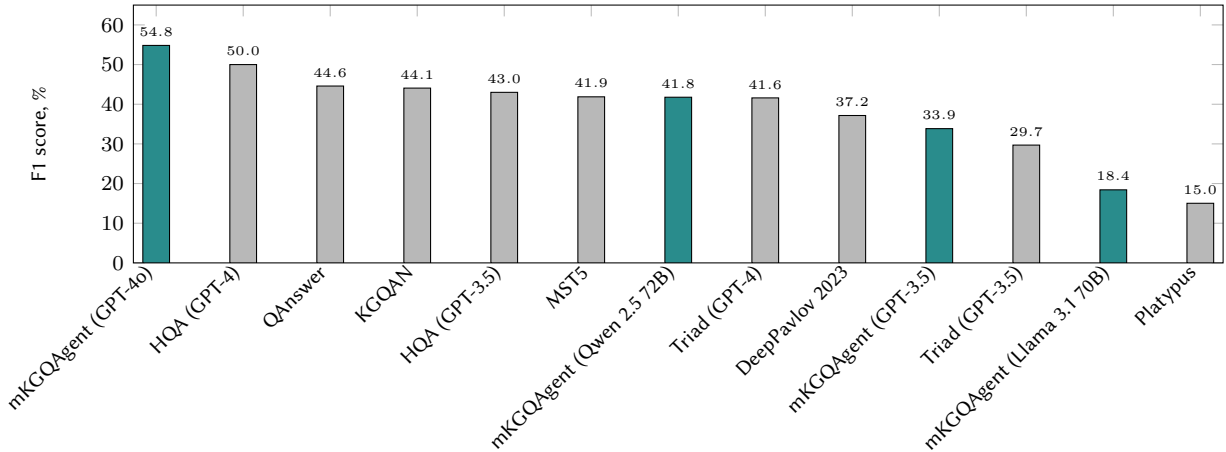
\begin{table*}[t!]
\centering
\caption{Evaluation results of \mKGQAgent (our approach) compared to the baselines that support multiple languages. They were conducted on the test subset of the \QALDplus. The best results per language are highlighted in bold.}
\resizebox{\textwidth}{!}{%
\begin{tabular}{lcccccccccc}
\toprule
 & \textbf{English} & \textbf{German} & \textbf{Spanish} & \textbf{French} & \textbf{Russian} & \textbf{Belarusian} & \textbf{Ukrainian} & \textbf{Lithuanian} & \textbf{Bashkir} & \textbf{Armenian} \\
 \midrule
\textbf{QAnswer} & 44.59 & 31.71 & 16.8 & \textbf{23.00} & 21.43 & \multicolumn{5}{c}{N/A} \\
\textbf{Platypus} & 15.03 & \multicolumn{2}{c}{N/A} & 4.17 & \multicolumn{6}{c}{N/A} \\
\textbf{DeepPavlov 2023} & 37.16 & \multicolumn{3}{c}{N/A} & 31.17 & \multicolumn{5}{c}{N/A} \\
\textbf{MST5} & 41.87 & 41.26 & N/A & 41.67 & \textbf{37.61} & 29.07 & \textbf{34.67} & \textbf{31.15} & 18.42 & N/A \\
\midrule
\textbf{mKGQAgent (GPT-3.5)} & 33.85 & 25.85 & 22.08 & 22.87 & 11.75 & 16.05 & 12.36 & 16.27 & 6.63 & 8.88 \\
\textbf{mKGQAgent (GPT-4o)} & \textbf{54.83} & \textbf{43.08} & \textbf{38.28} & 22.76 & 31.67 & \textbf{31.56} & 28.54 & 25.54 & \textbf{40.48} & 9.09 \\
\textbf{mKGQAgent (Qwen 2.5 72B)} & 41.77 & 21.86 & 17.96 & 17.60 & 12.92 & 10.81 & 10.86 & 9.56 & 10.40 & 0.00 \\
\textbf{mKGQAgent (Llama 3.1 70B)} & 18.42 & 19.64 & 5.23 & 17.70 & 5.22 & 5.73 & 6.70 & 9.52 & 4.65 & 15.07 \\
\bottomrule
\end{tabular}%
}
\label{tab:evaluation}

\end{table*}

The results presented in Figure \ref{fig:mkgqa-english-only} illustrate a comparative analysis between our \mKGQAgent approach (highlighted in teal) and various baseline methods (depicted in grey) on the English questions from the \QALDplus benchmark. 
Our \mKGQAgent (GPT-4o) achieves the highest F1 score of 54.83\%, surpassing all baselines, including HQA (GPT-4), which attains 50.00\%.
This demonstrates the effectiveness of our approach in leveraging structured planning and retrieval mechanisms to enhance semantic parsing performance.

Among the baselines, QAnswer (44.59\%) and KGQAN (44.07\%) show competitive results but still fall short of our top-performing model. 
Interestingly, HQA (GPT-3.5) achieves 43.00\%, indicating that the transition to GPT-4 has significantly improved query generation capabilities.
The performance of mKGQAgent (Qwen 2.5 72B) (41.87\%) and Triad (GPT-4) (41.77\%) suggests that large models, even with structured workflows, benefit from additional fine-tuning and experience pooling. 
Notably, our mKGQAgent (GPT-3.5) variant scores just 37.16\%, still outperforming several baselines but trailing behind its GPT-4o counterpart.



\subsection{Multilingual Comparison with the Baselines}
In Table \ref{tab:evaluation}, we present the evaluation results of our approach in comparison to the selected baselines that support multiple languages (see Section \ref{ssec:baselines}).

The experimental results demonstrate \mKGQAgent's robust performance across multiple languages on the \QALDplus benchmark. 
When implemented with GPT-4o, the system achieves state-of-the-art results with the F1 scores of 54.83\%, 43.08\%, 38.28\%, 31.56\%, and 40.48\%, respectively, for English, German, Spanish, Belarusian, and Bashkir. 
The languages using Cyrillic-based scripts (Russian, Belarusian, Ukrainian, and Bashkir) generally yield poorer results in comparison to the languages using Latin-based scripts.

While comparing \mKGQAgent to the other baselines, we see that QAnswer outperforms \mKGQAgent (GPT-3.5) on French; however, the difference is not substantial (23.00\% vs 22.87\%). 
The MST5 system significantly outperforms \mKGQAgent (GPT-4o) on Russian (37.61\% vs 31.67\%), Ukrainian (34.67\% vs 28.54\%), and Lithuanian (25.54\% vs 31.15\%).

\subsection{Machine Translation for non-English Questions}\label{ssec:mt_neamt}

The evaluation compares the performance of the models on native-language questions and those translated into English using machine translation (see Table \ref{tab:mt_neamt}).
\begin{table}[t]
\centering
\caption{Evaluation results of the \mKGQAgent (showing F1 score and percentages): A comparison between the quality of original language questions and the same questions translated into English using machine translation (MT).}
\label{tab:mt_neamt}
\resizebox{0.6\linewidth}{!}{%
\begin{tabular}{lcccccc}
\toprule
\multirow{2}{*}{} & \multicolumn{2}{c}{\textbf{German}} & \multicolumn{2}{c}{\textbf{Russian}} & \multicolumn{2}{c}{\textbf{Spanish}} \\ \cmidrule{2-7}
 & \textbf{Native} & \textbf{MT} & \textbf{Native} & \textbf{MT} & \textbf{Native} & \textbf{MT}  \\ \midrule
\textbf{mKGQAgent (GPT-3.5)} & 25.85 & 28.23 & 11.75 & 27.93 & 22.08 & 27.67  \\ 
\textbf{mKGQAgent (GPT-4o)} & 43.08 & 35.66 &  31.67 & 35.66 & 38.28 & 44.18 \\ 
\textbf{mKGQAgent (Qwen 2.5 72B)} & 21.86 & 30.95 &  12.92 & 32.97  & 17.96 & 31.35  \\ 
\textbf{mKGQAgent (Llama 3.1 70B)} & 19.64 & 17.29 &  5.22 & 17.29   & 5.23 & 9.63 \\
\bottomrule
\end{tabular}%
}
\end{table}
Across all models and languages, the performance of mKGQAgent is generally higher for translated questions than for native-language questions. 
This suggests that translating non-English questions into English before processing yields better results.

The \mKGQAgent based on GPT-4o achieves the highest performance across all settings, demonstrating the superior comprehension and reasoning capabilities of this model.
Qwen 2.5 72B exhibits strong performance in translation-based settings but falls behind GPT-4o.
The variance in performance between languages suggests that translation quality and linguistic characteristics play a role in how effectively mKGQAgent can process and answer questions.
In general, this comparison demonstrates that the translation of non-English questions into English consistently improves the quality and underlines the unequal quality distribution among the languages.

Table \ref{tab:neamt_mt_native} presents a comparative evaluation of the performance of MT against questions in their native language within the KGQA task. 
\begin{table}[t]
\centering
\caption{A comparison between the MT performance against the native questions (original language in the header row). We demonstrate the relative improvement when using MT (\gtriuparrow) or deterioration (\gtridownarrowred) in terms of F1 score.}
\label{tab:neamt_mt_native}
\resizebox{0.6\linewidth}{!}{%
\begin{tabular}{lrrr}
\toprule
\multirow{2}{*}{} & \multicolumn{1}{c}{\textbf{German}} & \multicolumn{1}{c}{\textbf{Russian}} & \multicolumn{1}{c}{\textbf{Spanish}} \\ \midrule
\textbf{mKGQAgent (GPT-3.5)} & +9.21\% \gtriuparrow & +137.69\% \gtriuparrow & +25.31\% \gtriuparrow \\ 
\textbf{mKGQAgent (GPT-4o)} & -17.22\% \gtridownarrowred & +12.59\% \gtriuparrow & +15.42\% \gtriuparrow \\ 
\textbf{mKGQAgent (Qwen 2.5 72B)} & +41.58\% \gtriuparrow & +155.21\% \gtriuparrow & +74.52\% \gtriuparrow \\ 
\textbf{mKGQAgent (Llama 3.1 70B)} & -11.97\% \gtridownarrowred & +231.46\% \gtriuparrow & +84.03\% \gtriuparrow \\ \bottomrule
\end{tabular}%
}
\end{table}
The results demonstrate that, for most models and languages MT yields improvements over native-language question answering. 
This effect is particularly pronounced in Russian and Spanish, where MT provides significant gains. 
GPT-4o, despite its strong overall performance, exhibits slight performance degradation when using MT  for German (-17.22\%), suggesting that this model may already be well optimized for handling German-language queries natively. 
Overall, these findings highlight the advantages of translations in multilingual KGQA, even when objectively strong LLMs are used.

\subsection{LLM Calls and Costs}\label{ssec:calls_costs}
An important aspect of using LLM agent frameworks is the number of model calls within one task solution, \ie in our case, we report the number of calls per generated \SPARQL query for an input question.
In addition, we report the estimated number of tokens per question and the underlying costs of the LLMs' usage.

\subsubsection{Costs calculation for the OpenAI models}

According to our calculations, \mKGQAgent \emph{requires 13.03 \LLM calls on average} to generate a \SPARQL query for an input question (\cf Table \ref{tab:impact}).
Consequently, every \LLM call consumes 144 input and 199 output tokens on average. 
This includes chat history that gradually grows during agent execution.
The pricing strategy of OpenAI is based on token consumption.
Therefore, we calculate the token-based price (TBP) as in Equation \ref{eq:tbp}.
\begin{equation}
\text{TBP} = \big[(n_{i} \times p_{i}) + (n_{o} \times p_{o})\big] \times n_{c} \times n_{q}
\label{eq:tbp}
\end{equation}
Where: $n_{i}$ represents the number of input tokens, $p_{i}$ represents the price per input token, $n_{o}$ represents the number of output tokens, $p_{o}$ represents the price per output token, $n_{c}$ represents the number of LLM calls per question, $n_{q}$ represents the number of questions.
This results in USD 0.48 per 100 questions for the GPT-3.5 and USD 3.06 per 100 questions for the GPT-4o, respectively (prices as of March 01, 2025).
For the costs of the different \SimpleAgent configurations, see Table \ref{tab:impact}.
\subsubsection{Costs calculation for the open source models}
For the open-source LLMs, we use the same values regarding the average number of \LLM calls to generate a \SPARQL query (13.03) and the average number of tokens per call--144 input tokens and 199 output tokens.
The pricing of open source models relies on the GPU hours of cloud providers and the model efficiency measured in tokens per second (tok/sec).
Therefore, we calculate the GPU hours-based price (GBP) as in Equation \ref{eq:gbp}.
\begin{equation}
    \text{GBP} = \frac{n_{q} \times n_{c} \times n_{o}}{r_\text{tok/sec}} \times p_{\text{GPU/sec}}
    \label{eq:gbp}
\end{equation}
Where: $r_\text{tok/sec}$ represents model efficiency rate (tok/sec), $p_\text{GPU/sec}$ price per GPU-second.
We estimated the market prices of our GPU experimental setup (2x Nvidia L40S GPU) according to one of the well-known cloud providers\footnote{\url{https://www.runpod.io/pricing}}.
The model performance (tok/sec) was retrieved from the official documentation of the respective models\footnote{Qwen: \url{https://qwen.readthedocs.io/en/latest/benchmark/speed_benchmark.html}, \\ Llama: \url{https://artificialanalysis.ai/models/llama-3-1-instruct-70b}} taking into account the usage of the vLLM framework for deployment and the size of the context window -- 16384 tokens.
Hence, for processing 100 questions, \mKGQAgent requires 1.96 GPU hours when using the Qwen model and 0.97 GPU hours when using the Llama model.
Therefore, the prices per 100 questions are USD 4.05 and 2.01, respectively.
For the costs of the different \SimpleAgent configurations, see Table \ref{tab:impact}.

\begin{table*}[t]
\centering
\setlength{\tabcolsep}{8pt}
\caption{Impact (regarding the baseline evaluation -- \SimpleAgent (Plan step + NEL tool) of individual components on the \QA quality (F1 score) measured on English questions of \QALDplus dataset. The strategy of pricing calculation is described in Section \ref{ssec:calls_costs}. \gtriuparrow~-- increases (the higher, the better), \gtriuparrowred~-- increases (the higher the worse).}

\label{tab:impact}
\resizebox{0.75\textwidth}{!}{%
\begin{tabular}{l|c|r|c|c|c}
\toprule
\textbf{Backbone LLM} & \textbf{F1 score, \%} & \textbf{Impact}\hspace{2ex} & \textbf{LLM calls}  & \textbf{Price per 100 questions} & \textbf{Price impact} \\ 
\midrule
\multicolumn{6}{c}{\SimpleAgent (Plan step + NEL tool)} \\
\midrule
GPT-3.5 & 23.15 & \multirow{4}{*}{N/A} \hspace{3ex} & \multirow{4}{*}{8.87 (avg.)} & USD 0.33 & \multirow{4}{*}{N/A}\\
GPT-4o & 34.37 &  & & USD 2.08 \\
Qwen 2.5 72B Instruct & 24.87 &  & &  USD 2.75 \\
Llama 3.1 70B Instruct & 8.12 &  & & USD 1.37 \\
\midrule
\multicolumn{6}{c}{\SimpleAgent + Experience Pool for Plan Step} \\
\midrule
GPT-3.5 & 31.17 &  +34.64\% \gtriuparrow & \multirow{4}{*}{9.71 (avg.)} & USD 0.36 & \multirow{4}{*}{+9.31\% \gtriuparrowred~(avg.)}\\
GPT-4o & 46.48 &   +35.23\% \gtriuparrow &  & USD 2.28 &\\
Qwen 2.5 72B Instruct & 29.45 &  +18.42\% \gtriuparrow & &   USD 3.02 &\\
Llama 3.1 70B Instruct & 17.06 &  +110.09\% \gtriuparrow  & & USD 1.49 &\\
\midrule
\multicolumn{6}{c}{\SimpleAgent + Experience Pool for Action Step} \\
\midrule
GPT-3.5 & 26.97 &  +16.50\% \gtriuparrow & \multirow{4}{*}{9.02 (avg.)} & USD 0.33 & \multirow{4}{*}{+1.73\% \gtriuparrowred~(avg.)}\\
GPT-4o & 52.68 &   +53.27\% \gtriuparrow & & USD 2.12 & \\
Qwen 2.5 72B Instruct & 32.87 &  +32.17\% \gtriuparrow &  & USD 2.80 & \\
Llama 3.1 70B Instruct & 15.58 &  +91.87\% \gtriuparrow &  & USD 1.39 & \\
\midrule
\multicolumn{6}{c}{\SimpleAgent + Feedback step} \\
\midrule
GPT-3.5 & 26.91 &  +16.24\% \gtriuparrow &  \multirow{4}{*}{10.93 (avg.)} & USD 0.40 & \multirow{4}{*}{+22.65\% \gtriuparrowred~(avg.)}\\
GPT-4o & 40.47 &  +17.74\% \gtriuparrow & & USD 2.57 & \\
Qwen 2.5 72B Instruct & 25.72 & +3.42\% \gtriuparrow & & USD 3.39 &\\
Llama 3.1 70B Instruct & 9.48 & +4.43\% \gtriuparrow & & USD 1.68 & \\
\midrule
\multicolumn{6}{c}{\mKGQAgent (Plan step + NEL tool + Experience Pool for Plan and Action + Feedback step)} \\
\midrule
GPT-3.5 & 33.85 &  +46.22\% \gtriuparrow& \multirow{4}{*}{13.03 (avg.)} & USD 0.48 & \multirow{4}{*}{+46.62\% \gtriuparrowred~(avg.)} \\
GPT-4o & 54.83 &  +59.53\% \gtriuparrow  & & USD 3.06 & \\
Qwen 2.5 72B Instruct & 41.77 & +67.95\% \gtriuparrow  &  & USD 4.05 & \\
Llama 3.1 70B Instruct & 20.42 & +151.47\% \gtriuparrow & & USD 2.01 & \\
\bottomrule
\end{tabular}%
}
\end{table*}
\subsection{Impact of Individual Components on the Quality (Ablation)}

To understand the contribution of each architectural component to the overall system performance, we conducted an ablation study using the English questions from the \QALDplus dataset. 
As our baseline system, we consider the \SimpleAgent with plan step and NEL tool components. 

The integration of the experience pool for the plan step demonstrated substantial improvements across all models. 
Notably, Llama exhibited the most substantial relative improvement of 110\%.

The addition of the experience pool to the action step proved particularly effective, especially when combined with GPT-4o, pushing the F1 score to 52.68\% (a 53.27\% improvement over baseline). 
This component's impact was consistently positive across all models.

The feedback step introduced more modest but still significant improvements. 
GPT-4o's performance with this component reached 40.47\%, representing a 17.74\% increase over the baseline. 

The full integration of all components in \mKGQAgent yielded the most impressive results, with GPT-4o achieving a peak F1 score of 54.83\%, marking a 59.53\% improvement over the baseline. 
This comprehensive integration demonstrated synergistic effects across all models, with even the initially lower-performing Llama showing a remarkable improvement of 151.47\% (see Table \ref{tab:impact}). 
These results strongly indicate that the combination of all components creates a more robust and effective KGQA system.

\section{Discussion and Research Questions}\label{sec:discussion}



\begin{description}[labelindent=-2pt]
    \item[\RQ{1}] The analysis reveals that the proposed agent architecture enables more accurate \SPARQL query generation. 
In particular, \mKGQAgent achieved state-of-the-art results (54.83\% F1 score) on English and demonstrated superior quality on German, Spanish, Belarusian, and Bashkir.

    \item[\RQ{2}] The evaluation indicates that the full \mKGQAgent setup with all components achieved substantially better quality but requires additional computational resources.
For example, the \SimpleAgent requires 8.87 LLM calls on average to achieve the end goal, while the final \mKGQAgent requires 13.03 LLM calls on average. 

    \item[\RQ{3}] Our work indicates that multilingual \SPARQL generation presents significant challenges even to the state-of-the-art LLMs. 
In particular, even among European languages, the quality of SPARQL query generation may degrade by more than a factor of three.

   \item[\RQ{4}] Our results indicate that machine translation generally leads to higher KGQA performance compared to processing questions in their native languages. 
However, the effectiveness of translation-based approaches varies by language and model. 
\end{description}

\section{Conclusion}\label{sec:conclusion}
The paper introduces a novel \LLM agent framework called \mKGQAgent for the multilingual Text-to-\SPARQL task. 
The experiments carried out have shown that each step of the \mKGQAgent workflow contributes positively to the quality of the results. 
The \mKGQAgent substantially outperforms previous systems in the English, German, Spanish, Belarusian, and Bashkir questions of the \QALDplus data set.

We highlighted significant challenges when LLMs deal with non-English languages, especially low-resource ones.
The latter challenge can be partially covered by the use of MT  techniques, which was demonstrated by our experiments.
However, the use of different translation techniques requires further systematic study to identify settings where each of them performs best.
Despite this, the \mKGQAgent framework demonstrates a promising approach to \KGQA by adopting the LLM agent paradigm. 
While it shows its ability to work with multiple languages having reasonably good quality, we also demonstrated the trade-off between the quality and computational costs that increase with the agent paradigm adoption. 

\begin{acknowledgments}
  This work has been partially supported by grants for the ITZBund\footnote{\url{https://www.itzbund.de/}}-funded research project \qq{QA4CB---Entwicklung von Question-Answering-Komponenten zur Erweiterung des Chatbot-Frameworks} at the Leipzig University of Applied Sciences in Leipzig (Germany).
\end{acknowledgments}

\section*{Declaration on Generative AI}
 During the preparation of this work, the authors used ChatGPT by OpenAI in order to: Grammar and spelling check.
 After using this service, the authors reviewed and edited the content as needed and take(s) full responsibility for the publication’s content. 

\bibliography{my}

\begin{thebibliography}{37}
\expandafter\ifx\csname natexlab\endcsname\relax\def\natexlab#1{#1}\fi
\providecommand{\url}[1]{\texttt{#1}}
\providecommand{\href}[2]{#2}
\providecommand{\path}[1]{#1}
\providecommand{\DOIprefix}{doi:}
\providecommand{\ArXivprefix}{arXiv:}
\providecommand{\URLprefix}{URL: }
\providecommand{\Pubmedprefix}{pmid:}
\providecommand{\doi}[1]{\href{http://dx.doi.org/#1}{\path{#1}}}
\providecommand{\Pubmed}[1]{\href{pmid:#1}{\path{#1}}}
\providecommand{\bibinfo}[2]{#2}
\ifx\xfnm\relax \def\xfnm[#1]{\unskip,\space#1}\fi
\bibitem[{Diefenbach et~al.(2019)Diefenbach, Migliatti, Qawasmeh, Lully, Singh, and Maret}]{qanswer}
\bibinfo{author}{D.~Diefenbach}, \bibinfo{author}{P.~H. Migliatti}, \bibinfo{author}{O.~Qawasmeh}, \bibinfo{author}{V.~Lully}, \bibinfo{author}{K.~Singh}, \bibinfo{author}{P.~Maret},
\newblock \bibinfo{title}{{QAnswer}: A question answering prototype bridging the gap between a considerable part of the lod cloud and end-users},
\newblock in: \bibinfo{booktitle}{The World Wide Web Conference}, WWW '19, \bibinfo{publisher}{Association for Computing Machinery}, \bibinfo{address}{New York, NY, USA}, \bibinfo{year}{2019}, p. \bibinfo{pages}{3507–3510}. \DOIprefix\doi{10.1145/3308558.3314124}.
\bibitem[{Turganbay et~al.(2023)Turganbay, Surkov, Evseev, and Drobyshevskiy}]{deeppavlov2023}
\bibinfo{author}{R.~Turganbay}, \bibinfo{author}{V.~Surkov}, \bibinfo{author}{D.~Evseev}, \bibinfo{author}{M.~Drobyshevskiy},
\newblock \bibinfo{title}{Generative question answering systems over knowledge graphs and text},
\newblock volume~\bibinfo{volume}{22}, \bibinfo{year}{2023}, pp. \bibinfo{pages}{1112--1126}. \DOIprefix\doi{10.28995/2075-7182-2023-22-1112-1126}.
\bibitem[{Srivastava et~al.(2024)Srivastava, Ma, Vollmers, Zahera, Moussallem, and Ngomo}]{srivastava2024mst5}
\bibinfo{author}{N.~Srivastava}, \bibinfo{author}{M.~Ma}, \bibinfo{author}{D.~Vollmers}, \bibinfo{author}{H.~Zahera}, \bibinfo{author}{D.~Moussallem}, \bibinfo{author}{A.-C.~N. Ngomo},
\newblock \bibinfo{title}{{MST5}--multilingual question answering over knowledge graphs},
\newblock \bibinfo{journal}{arXiv preprint arXiv:2407.06041}  (\bibinfo{year}{2024}).
\bibitem[{Jiang et~al.(2024)Jiang, Zhou, Zhao, Song, Zhu, Zhu, and Wen}]{jiang2024kg}
\bibinfo{author}{J.~Jiang}, \bibinfo{author}{K.~Zhou}, \bibinfo{author}{W.~X. Zhao}, \bibinfo{author}{Y.~Song}, \bibinfo{author}{C.~Zhu}, \bibinfo{author}{H.~Zhu}, \bibinfo{author}{J.-R. Wen},
\newblock \bibinfo{title}{{KG-Agent}: An efficient autonomous agent framework for complex reasoning over knowledge graph},
\newblock \bibinfo{journal}{arXiv preprint arXiv:2402.11163}  (\bibinfo{year}{2024}).
\bibitem[{Huang et~al.(2024)Huang, Cheng, Huang, Shen, Xu, Zhang, and Qu}]{huang2024queryagent}
\bibinfo{author}{X.~Huang}, \bibinfo{author}{S.~Cheng}, \bibinfo{author}{S.~Huang}, \bibinfo{author}{J.~Shen}, \bibinfo{author}{Y.~Xu}, \bibinfo{author}{C.~Zhang}, \bibinfo{author}{Y.~Qu},
\newblock \bibinfo{title}{{QueryAgent}: A reliable and efficient reasoning framework with environmental feedback based self-correction},
\newblock \bibinfo{journal}{arXiv preprint arXiv:2403.11886}  (\bibinfo{year}{2024}).
\bibitem[{Li et~al.(2023)Li, Zhang, and Sun}]{li2023metaagents}
\bibinfo{author}{Y.~Li}, \bibinfo{author}{Y.~Zhang}, \bibinfo{author}{L.~Sun},
\newblock \bibinfo{title}{{MetaAgents}: Simulating interactions of human behaviors for {LLM-based} task-oriented coordination via collaborative generative agents},
\newblock \bibinfo{journal}{CoRR} \bibinfo{volume}{abs/2310.06500} (\bibinfo{year}{2023}). \DOIprefix\doi{10.48550/ARXIV.2310.06500}. \href{http://arxiv.org/abs/2310.06500}{{\tt arXiv:2310.06500}}.
\bibitem[{Diefenbach et~al.(2017)Diefenbach, Singh, Both, Cherix, Lange, and Auer}]{diefenbach2017qanaryecosystem}
\bibinfo{author}{D.~Diefenbach}, \bibinfo{author}{K.~Singh}, \bibinfo{author}{A.~Both}, \bibinfo{author}{D.~Cherix}, \bibinfo{author}{C.~Lange}, \bibinfo{author}{S.~Auer},
\newblock \bibinfo{title}{The {Qanary} ecosystem: Getting new insights by composing question answering pipelines},
\newblock in: \bibinfo{editor}{J.~Cabot}, \bibinfo{editor}{R.~De~Virgilio}, \bibinfo{editor}{R.~Torlone} (Eds.), \bibinfo{booktitle}{Web Engineering}, \bibinfo{publisher}{Springer International Publishing}, \bibinfo{address}{Cham}, \bibinfo{year}{2017}, pp. \bibinfo{pages}{171--189}.
\bibitem[{Correa et~al.(2020)Correa, Ho, Callaway, and Griffiths}]{correa2020resource}
\bibinfo{author}{C.~G. Correa}, \bibinfo{author}{M.~K. Ho}, \bibinfo{author}{F.~Callaway}, \bibinfo{author}{T.~L. Griffiths},
\newblock \bibinfo{title}{Resource-rational task decomposition to minimize planning costs},
\newblock in: \bibinfo{booktitle}{Proceedings of the 42th Annual Meeting of the Cognitive Science Society - Developing a Mind: Learning in Humans, Animals, and Machines, CogSci 2020, virtual, July 29 - August 1, 2020}, \bibinfo{publisher}{cognitivesciencesociety.org}, \bibinfo{year}{2020}. \URLprefix \url{https://cogsci.mindmodeling.org/2020/papers/0746/index.html}.
\bibitem[{Perevalov et~al.(2022{\natexlab{a}})Perevalov, Diefenbach, Usbeck, and Both}]{perevalov2022qald}
\bibinfo{author}{A.~Perevalov}, \bibinfo{author}{D.~Diefenbach}, \bibinfo{author}{R.~Usbeck}, \bibinfo{author}{A.~Both},
\newblock \bibinfo{title}{{QALD-9-plus}: A multilingual dataset for question answering over {DBpedia} and {Wikidata} translated by native speakers},
\newblock in: \bibinfo{booktitle}{2022 IEEE 16th International Conference on Semantic Computing (ICSC)}, \bibinfo{organization}{IEEE}, \bibinfo{year}{2022}{\natexlab{a}}, pp. \bibinfo{pages}{229--234}.
\bibitem[{Perevalov et~al.(2022{\natexlab{b}})Perevalov, Yan, Kovriguina, Jiang, Both, and Usbeck}]{leaderboard}
\bibinfo{author}{A.~Perevalov}, \bibinfo{author}{X.~Yan}, \bibinfo{author}{L.~Kovriguina}, \bibinfo{author}{L.~Jiang}, \bibinfo{author}{A.~Both}, \bibinfo{author}{R.~Usbeck},
\newblock \bibinfo{title}{Knowledge graph question answering leaderboard: A community resource to prevent a replication crisis},
\newblock in: \bibinfo{booktitle}{Proceedings of the Language Resources and Evaluation Conference}, \bibinfo{publisher}{European Language Resources Association}, \bibinfo{address}{Marseille, France}, \bibinfo{year}{2022}{\natexlab{b}}, pp. \bibinfo{pages}{2998--3007}. \URLprefix \url{https://aclanthology.org/2022.lrec-1.321}.
\bibitem[{Perevalov et~al.(2024)Perevalov, Both, and Ngonga~Ngomo}]{perevalovmultilingual}
\bibinfo{author}{A.~Perevalov}, \bibinfo{author}{A.~Both}, \bibinfo{author}{A.-C. Ngonga~Ngomo},
\newblock \bibinfo{title}{Multilingual question answering systems for knowledge graphs---a survey},
\newblock \bibinfo{journal}{Semantic Web} \bibinfo{volume}{15} (\bibinfo{year}{2024}) \bibinfo{pages}{2089--2124}.
\bibitem[{Punjani et~al.(2018)Punjani, Singh, Both, Koubarakis, Angelidis, Bereta, Beris, Bilidas, Ioannidis, Karalis, Lange, Pantazi, Papaloukas, and Stamoulis}]{10.1145/3281354.3281362}
\bibinfo{author}{D.~Punjani}, \bibinfo{author}{K.~Singh}, \bibinfo{author}{A.~Both}, \bibinfo{author}{M.~Koubarakis}, \bibinfo{author}{I.~Angelidis}, \bibinfo{author}{K.~Bereta}, \bibinfo{author}{T.~Beris}, \bibinfo{author}{D.~Bilidas}, \bibinfo{author}{T.~Ioannidis}, \bibinfo{author}{N.~Karalis}, \bibinfo{author}{C.~Lange}, \bibinfo{author}{D.~Pantazi}, \bibinfo{author}{C.~Papaloukas}, \bibinfo{author}{G.~Stamoulis},
\newblock \bibinfo{title}{Template-based question answering over linked geospatial data},
\newblock in: \bibinfo{booktitle}{Proceedings of the 12th Workshop on Geographic Information Retrieval}, GIR'18, \bibinfo{publisher}{Association for Computing Machinery}, \bibinfo{address}{New York, NY, USA}, \bibinfo{year}{2018}. \URLprefix \url{https://doi.org/10.1145/3281354.3281362}. \DOIprefix\doi{10.1145/3281354.3281362}.
\bibitem[{Pellissier~Tanon et~al.(2018)Pellissier~Tanon, de~Assun{\c{c}}{\~a}o, Caron, and Suchanek}]{platypus}
\bibinfo{author}{T.~Pellissier~Tanon}, \bibinfo{author}{M.~D. de~Assun{\c{c}}{\~a}o}, \bibinfo{author}{E.~Caron}, \bibinfo{author}{F.~M. Suchanek},
\newblock \bibinfo{title}{Demoing {Platypus} -- a multilingual question answering platform for {Wikidata}},
\newblock in: \bibinfo{editor}{A.~Gangemi}, \bibinfo{editor}{A.~L. Gentile}, \bibinfo{editor}{A.~G. Nuzzolese}, \bibinfo{editor}{S.~Rudolph}, \bibinfo{editor}{M.~Maleshkova}, \bibinfo{editor}{H.~Paulheim}, \bibinfo{editor}{J.~Z. Pan}, \bibinfo{editor}{M.~Alam} (Eds.), \bibinfo{booktitle}{The Semantic Web: ESWC 2018 Satellite Events}, \bibinfo{publisher}{Springer International Publishing}, \bibinfo{address}{Cham}, \bibinfo{year}{2018}, pp. \bibinfo{pages}{111--116}.
\bibitem[{Omar et~al.(2023)Omar, Dhall, Kalnis, and Mansour}]{omar2023universal}
\bibinfo{author}{R.~Omar}, \bibinfo{author}{I.~Dhall}, \bibinfo{author}{P.~Kalnis}, \bibinfo{author}{E.~Mansour},
\newblock \bibinfo{title}{A universal question-answering platform for knowledge graphs},
\newblock \bibinfo{journal}{Proceedings of the ACM on Management of Data} \bibinfo{volume}{1} (\bibinfo{year}{2023}) \bibinfo{pages}{1--25}.
\bibitem[{Zhou et~al.(2021)Zhou, Geng, Shen, Zhang, and Jiang}]{zhou-etal-2021-improving}
\bibinfo{author}{Y.~Zhou}, \bibinfo{author}{X.~Geng}, \bibinfo{author}{T.~Shen}, \bibinfo{author}{W.~Zhang}, \bibinfo{author}{D.~Jiang},
\newblock \bibinfo{title}{Improving zero-shot cross-lingual transfer for multilingual question answering over knowledge graph},
\newblock in: \bibinfo{editor}{K.~Toutanova}, \bibinfo{editor}{A.~Rumshisky}, \bibinfo{editor}{L.~Zettlemoyer}, \bibinfo{editor}{D.~Hakkani-Tur}, \bibinfo{editor}{I.~Beltagy}, \bibinfo{editor}{S.~Bethard}, \bibinfo{editor}{R.~Cotterell}, \bibinfo{editor}{T.~Chakraborty}, \bibinfo{editor}{Y.~Zhou} (Eds.), \bibinfo{booktitle}{Proceedings of the 2021 Conference of the North American Chapter of the Association for Computational Linguistics: Human Language Technologies}, \bibinfo{publisher}{Association for Computational Linguistics}, \bibinfo{address}{Online}, \bibinfo{year}{2021}, pp. \bibinfo{pages}{5822--5834}. \URLprefix \url{https://aclanthology.org/2021.naacl-main.465/}. \DOIprefix\doi{10.18653/v1/2021.naacl-main.465}.
\bibitem[{Zhang et~al.(2023)Zhang, Wang, Wang, and Zhang}]{zhang-etal-2023-xsemplr}
\bibinfo{author}{Y.~Zhang}, \bibinfo{author}{J.~Wang}, \bibinfo{author}{Z.~Wang}, \bibinfo{author}{R.~Zhang},
\newblock \bibinfo{title}{{XS}em{PLR}: Cross-lingual semantic parsing in multiple natural languages and meaning representations},
\newblock in: \bibinfo{booktitle}{Proceedings of the 61st Annual Meeting of the Association for Computational Linguistics (Volume 1: Long Papers)}, \bibinfo{publisher}{Association for Computational Linguistics}, \bibinfo{address}{Toronto, Canada}, \bibinfo{year}{2023}, pp. \bibinfo{pages}{15918--15947}. \URLprefix \url{https://aclanthology.org/2023.acl-long.887}.
\bibitem[{Tan et~al.(2023)Tan, Zhang, Chen, Ali, Hua, and Qi}]{TAN2023120721}
\bibinfo{author}{Y.~Tan}, \bibinfo{author}{X.~Zhang}, \bibinfo{author}{Y.~Chen}, \bibinfo{author}{Z.~Ali}, \bibinfo{author}{Y.~Hua}, \bibinfo{author}{G.~Qi},
\newblock \bibinfo{title}{{CLRN}: A reasoning network for multi-relation question answering over cross-lingual knowledge graphs},
\newblock \bibinfo{journal}{Expert Systems with Applications} \bibinfo{volume}{231} (\bibinfo{year}{2023}) \bibinfo{pages}{120721}. \URLprefix \url{https://www.sciencedirect.com/science/article/pii/S095741742301223X}. \DOIprefix\doi{https://doi.org/10.1016/j.eswa.2023.120721}.
\bibitem[{Zong et~al.(2024)Zong, Yan, Lu, Shao, Huang, Chang, and Zhuang}]{zong2024triad}
\bibinfo{author}{C.~Zong}, \bibinfo{author}{Y.~Yan}, \bibinfo{author}{W.~Lu}, \bibinfo{author}{J.~Shao}, \bibinfo{author}{Y.~Huang}, \bibinfo{author}{H.~Chang}, \bibinfo{author}{Y.~Zhuang},
\newblock \bibinfo{title}{Triad: A framework leveraging a multi-role {LLM}-based agent to solve knowledge base question answering},
\newblock in: \bibinfo{booktitle}{Proceedings of the 2024 Conference on Empirical Methods in Natural Language Processing}, \bibinfo{year}{2024}, pp. \bibinfo{pages}{1698--1710}.
\bibitem[{Lehmann et~al.(2024)Lehmann, Bhandiwad, Gattogi, and Vahdati}]{lehmann2024beyond}
\bibinfo{author}{J.~Lehmann}, \bibinfo{author}{D.~Bhandiwad}, \bibinfo{author}{P.~Gattogi}, \bibinfo{author}{S.~Vahdati},
\newblock \bibinfo{title}{Beyond boundaries: A human-like approach for question answering over structured and unstructured information sources},
\newblock \bibinfo{journal}{Transactions of the Association for Computational Linguistics} \bibinfo{volume}{12} (\bibinfo{year}{2024}) \bibinfo{pages}{786--802}.
\bibitem[{Xiong et~al.(2024)Xiong, Bao, and Zhao}]{xiong2024interactive}
\bibinfo{author}{G.~Xiong}, \bibinfo{author}{J.~Bao}, \bibinfo{author}{W.~Zhao},
\newblock \bibinfo{title}{{Interactive-KBQA}: Multi-turn interactions for knowledge base question answering with large language models},
\newblock \bibinfo{journal}{CoRR} \bibinfo{volume}{abs/2402.15131} (\bibinfo{year}{2024}). \DOIprefix\doi{10.48550/ARXIV.2402.15131}.
\bibitem[{Mialon et~al.(2023)Mialon, Dessi, Lomeli, Nalmpantis, Pasunuru, Raileanu, Roziere, Schick, Dwivedi-Yu, Celikyilmaz et~al.}]{mialon2023augmented}
\bibinfo{author}{G.~Mialon}, \bibinfo{author}{R.~Dessi}, \bibinfo{author}{M.~Lomeli}, \bibinfo{author}{C.~Nalmpantis}, \bibinfo{author}{R.~Pasunuru}, \bibinfo{author}{R.~Raileanu}, \bibinfo{author}{B.~Roziere}, \bibinfo{author}{T.~Schick}, \bibinfo{author}{J.~Dwivedi-Yu}, \bibinfo{author}{A.~Celikyilmaz}, et~al.,
\newblock \bibinfo{title}{Augmented language models: a survey},
\newblock \bibinfo{journal}{Transactions on Machine Learning Research}  (\bibinfo{year}{2023}).
\bibitem[{Wang et~al.(2024)Wang, Ma, Feng, Zhang, Yang, Zhang, Chen, Tang, Chen, Lin et~al.}]{wang2024survey}
\bibinfo{author}{L.~Wang}, \bibinfo{author}{C.~Ma}, \bibinfo{author}{X.~Feng}, \bibinfo{author}{Z.~Zhang}, \bibinfo{author}{H.~Yang}, \bibinfo{author}{J.~Zhang}, \bibinfo{author}{Z.~Chen}, \bibinfo{author}{J.~Tang}, \bibinfo{author}{X.~Chen}, \bibinfo{author}{Y.~Lin}, et~al.,
\newblock \bibinfo{title}{A survey on large language model based autonomous agents},
\newblock \bibinfo{journal}{Frontiers of Computer Science} \bibinfo{volume}{18} (\bibinfo{year}{2024}) \bibinfo{pages}{186345}. \DOIprefix\doi{10.1007/s11704-024-40231-1}.
\bibitem[{Luo et~al.(2023)Luo, Yang, Meng, Li, Zhou, and Zhang}]{luo2023empirical}
\bibinfo{author}{Y.~Luo}, \bibinfo{author}{Z.~Yang}, \bibinfo{author}{F.~Meng}, \bibinfo{author}{Y.~Li}, \bibinfo{author}{J.~Zhou}, \bibinfo{author}{Y.~Zhang},
\newblock \bibinfo{title}{An empirical study of catastrophic forgetting in large language models during continual fine-tuning},
\newblock \bibinfo{journal}{arXiv preprint arXiv:2308.08747}  (\bibinfo{year}{2023}).
\bibitem[{Vrande{\v{c}}i{\'c} and Kr{\"o}tzsch(2014)}]{vrandevcic2014wikidata}
\bibinfo{author}{D.~Vrande{\v{c}}i{\'c}}, \bibinfo{author}{M.~Kr{\"o}tzsch},
\newblock \bibinfo{title}{Wikidata: a free collaborative knowledgebase},
\newblock \bibinfo{journal}{Communications of the ACM} \bibinfo{volume}{57} (\bibinfo{year}{2014}) \bibinfo{pages}{78--85}. \DOIprefix\doi{10.1145/2629489}.
\bibitem[{Huys et~al.(2015)Huys, Lally, Faulkner, Eshel, Seifritz, Gershman, Dayan, and Roiser}]{huys2015interplay}
\bibinfo{author}{Q.~J. Huys}, \bibinfo{author}{N.~Lally}, \bibinfo{author}{P.~Faulkner}, \bibinfo{author}{N.~Eshel}, \bibinfo{author}{E.~Seifritz}, \bibinfo{author}{S.~J. Gershman}, \bibinfo{author}{P.~Dayan}, \bibinfo{author}{J.~P. Roiser},
\newblock \bibinfo{title}{Interplay of approximate planning strategies},
\newblock \bibinfo{journal}{Proceedings of the National Academy of Sciences} \bibinfo{volume}{112} (\bibinfo{year}{2015}) \bibinfo{pages}{3098--3103}. \DOIprefix\doi{10.1073/pnas.1414219112}.
\bibitem[{Auer et~al.(2007)Auer, Bizer, Kobilarov, Lehmann, Cyganiak, and Ives}]{dbpedia}
\bibinfo{author}{S.~Auer}, \bibinfo{author}{C.~Bizer}, \bibinfo{author}{G.~Kobilarov}, \bibinfo{author}{J.~Lehmann}, \bibinfo{author}{R.~Cyganiak}, \bibinfo{author}{Z.~Ives},
\newblock \bibinfo{title}{{DBpedia}: A nucleus for a web of open data},
\newblock in: \bibinfo{booktitle}{The semantic web}, \bibinfo{publisher}{Springer}, \bibinfo{year}{2007}, pp. \bibinfo{pages}{722--735}.
\bibitem[{Vrande\v{c}i\'{c} and Kr\"{o}tzsch(2014)}]{Wikidata}
\bibinfo{author}{D.~Vrande\v{c}i\'{c}}, \bibinfo{author}{M.~Kr\"{o}tzsch},
\newblock \bibinfo{title}{Wikidata: A free collaborative knowledgebase},
\newblock \bibinfo{journal}{Commun. ACM} \bibinfo{volume}{57} (\bibinfo{year}{2014}) \bibinfo{pages}{78–85}. \DOIprefix\doi{10.1145/2629489}.
\bibitem[{Soruco et~al.(2023)Soruco, Collarana, Both, and Usbeck}]{soruco2023qald}
\bibinfo{author}{J.~Soruco}, \bibinfo{author}{D.~Collarana}, \bibinfo{author}{A.~Both}, \bibinfo{author}{R.~Usbeck},
\newblock \bibinfo{title}{{QALD-9-ES}: A {Spanish} dataset for question answering systems},
\newblock in: \bibinfo{booktitle}{Knowledge Graphs: Semantics, Machine Learning, and Languages}, \bibinfo{publisher}{IOS Press}, \bibinfo{year}{2023}, pp. \bibinfo{pages}{38--52}.
\bibitem[{Usbeck et~al.(2019)Usbeck, R{\"o}der, Hoffmann, Conrads, Huthmann, Ngonga-Ngomo, Demmler, and Unger}]{usbeck2019benchmarking}
\bibinfo{author}{R.~Usbeck}, \bibinfo{author}{M.~R{\"o}der}, \bibinfo{author}{M.~Hoffmann}, \bibinfo{author}{F.~Conrads}, \bibinfo{author}{J.~Huthmann}, \bibinfo{author}{A.-C. Ngonga-Ngomo}, \bibinfo{author}{C.~Demmler}, \bibinfo{author}{C.~Unger},
\newblock \bibinfo{title}{Benchmarking question answering systems},
\newblock \bibinfo{journal}{Semantic Web} \bibinfo{volume}{10} (\bibinfo{year}{2019}) \bibinfo{pages}{293--304}.
\bibitem[{Lin et~al.(2024)Lin, Tang, Tang, Yang, Chen, Wang, Xiao, Dang, Gan, and Han}]{MLSYS2024_42a452cb}
\bibinfo{author}{J.~Lin}, \bibinfo{author}{J.~Tang}, \bibinfo{author}{H.~Tang}, \bibinfo{author}{S.~Yang}, \bibinfo{author}{W.-M. Chen}, \bibinfo{author}{W.-C. Wang}, \bibinfo{author}{G.~Xiao}, \bibinfo{author}{X.~Dang}, \bibinfo{author}{C.~Gan}, \bibinfo{author}{S.~Han},
\newblock \bibinfo{title}{{AWQ}: Activation-aware weight quantization for on-device {LLM} compression and acceleration},
\newblock in: \bibinfo{editor}{P.~Gibbons}, \bibinfo{editor}{G.~Pekhimenko}, \bibinfo{editor}{C.~D. Sa} (Eds.), \bibinfo{booktitle}{Proceedings of Machine Learning and Systems}, volume~\bibinfo{volume}{6}, \bibinfo{year}{2024}, pp. \bibinfo{pages}{87--100}. \URLprefix \url{https://proceedings.mlsys.org/paper_files/paper/2024/file/42a452cbafa9dd64e9ba4aa95cc1ef21-Paper-Conference.pdf}.
\bibitem[{Kwon et~al.(2023)Kwon, Li, Zhuang, Sheng, Zheng, Yu, Gonzalez, Zhang, and Stoica}]{kwon2023efficient}
\bibinfo{author}{W.~Kwon}, \bibinfo{author}{Z.~Li}, \bibinfo{author}{S.~Zhuang}, \bibinfo{author}{Y.~Sheng}, \bibinfo{author}{L.~Zheng}, \bibinfo{author}{C.~H. Yu}, \bibinfo{author}{J.~E. Gonzalez}, \bibinfo{author}{H.~Zhang}, \bibinfo{author}{I.~Stoica},
\newblock \bibinfo{title}{Efficient memory management for large language model serving with pagedattention},
\newblock in: \bibinfo{booktitle}{Proceedings of the ACM SIGOPS 29th Symposium on Operating Systems Principles}, \bibinfo{year}{2023}.
\bibitem[{Wang et~al.(2024)Wang, Yang, Huang, Yang, Majumder, and Wei}]{wang2024multilingual}
\bibinfo{author}{L.~Wang}, \bibinfo{author}{N.~Yang}, \bibinfo{author}{X.~Huang}, \bibinfo{author}{L.~Yang}, \bibinfo{author}{R.~Majumder}, \bibinfo{author}{F.~Wei},
\newblock \bibinfo{title}{Multilingual {E5} text embeddings: A technical report},
\newblock \bibinfo{journal}{arXiv preprint arXiv:2402.05672}  (\bibinfo{year}{2024}).
\bibitem[{Muennighoff et~al.(2022)Muennighoff, Tazi, Magne, and Reimers}]{muennighoff2022mteb}
\bibinfo{author}{N.~Muennighoff}, \bibinfo{author}{N.~Tazi}, \bibinfo{author}{L.~Magne}, \bibinfo{author}{N.~Reimers},
\newblock \bibinfo{title}{{MTEB}: Massive text embedding benchmark},
\newblock \bibinfo{journal}{arXiv preprint arXiv:2210.07316}  (\bibinfo{year}{2022}).
\bibitem[{Sakor et~al.(2020)Sakor, Singh, Patel, and Vidal}]{falcon2}
\bibinfo{author}{A.~Sakor}, \bibinfo{author}{K.~Singh}, \bibinfo{author}{A.~Patel}, \bibinfo{author}{M.-E. Vidal},
\newblock \bibinfo{title}{Falcon 2.0: An entity and relation linking tool over {Wikidata}},
\newblock in: \bibinfo{booktitle}{Proceedings of the 29th ACM International Conference on Information \& Knowledge Management}, CIKM '20, \bibinfo{publisher}{Association for Computing Machinery}, \bibinfo{address}{New York, NY, USA}, \bibinfo{year}{2020}, p. \bibinfo{pages}{3141–3148}. \DOIprefix\doi{10.1145/3340531.3412777}.
\bibitem[{Perevalov et~al.(2022)Perevalov, Both, Diefenbach, and Ngonga~Ngomo}]{perevalovMt}
\bibinfo{author}{A.~Perevalov}, \bibinfo{author}{A.~Both}, \bibinfo{author}{D.~Diefenbach}, \bibinfo{author}{A.-C. Ngonga~Ngomo},
\newblock \bibinfo{title}{Can machine translation be a reasonable alternative for multilingual question answering systems over knowledge graphs?},
\newblock in: \bibinfo{booktitle}{Proceedings of the ACM Web Conference 2022}, WWW '22, \bibinfo{publisher}{Association for Computing Machinery}, \bibinfo{address}{New York, NY, USA}, \bibinfo{year}{2022}, p. \bibinfo{pages}{977–986}. \URLprefix \url{https://doi.org/10.1145/3485447.3511940}. \DOIprefix\doi{10.1145/3485447.3511940}.
\bibitem[{Srivastava et~al.(2023)Srivastava, Perevalov, Kuchelev, Moussallem, Ngonga~Ngomo, and Both}]{linguaFranca}
\bibinfo{author}{N.~Srivastava}, \bibinfo{author}{A.~Perevalov}, \bibinfo{author}{D.~Kuchelev}, \bibinfo{author}{D.~Moussallem}, \bibinfo{author}{A.-C. Ngonga~Ngomo}, \bibinfo{author}{A.~Both},
\newblock \bibinfo{title}{Lingua franca – entity-aware machine translation approach for question answering over knowledge graphs},
\newblock in: \bibinfo{booktitle}{Proceedings of the 12th Knowledge Capture Conference 2023}, K-CAP '23, \bibinfo{publisher}{Association for Computing Machinery}, \bibinfo{address}{New York, NY, USA}, \bibinfo{year}{2023}, p. \bibinfo{pages}{122–130}. \URLprefix \url{https://doi.org/10.1145/3587259.3627567}. \DOIprefix\doi{10.1145/3587259.3627567}.
\bibitem[{Tiedemann and Thottingal(2020)}]{helsinkiNLP}
\bibinfo{author}{J.~Tiedemann}, \bibinfo{author}{S.~Thottingal},
\newblock \bibinfo{title}{{OPUS-MT} — {B}uilding open translation services for the {W}orld},
\newblock in: \bibinfo{booktitle}{Proceedings of the 22nd Annual Conference of the European Association for Machine Translation (EAMT)}, \bibinfo{address}{Lisbon, Portugal}, \bibinfo{year}{2020}.

\end{thebibliography}



\end{document}